\documentclass[final,5p,times,twocolumn]{elsarticle}

\usepackage{amssymb}
\usepackage{amsmath}
\usepackage{diagbox}
\biboptions{numbers,sort&compress}
\usepackage[colorlinks]{hyperref}
\usepackage{caption}
\usepackage{multirow}
\usepackage[below]{placeins}

\begin{document}
\captionsetup[figure]{labelfont={bf},labelformat={default},labelsep=period,name={Fig.}}
\begin{frontmatter}


\title{Component-aware anomaly detection framework for adjustable and logical industrial visual inspection}


\author[Address1]{Tongkun Liu}
\author[Address1,Address2]{Bing Li}
\author[Address1]{Xiao Du}
\author[Address1]{Bingke Jiang}
\author[Address1]{Xiao Jin}
\author[Address1]{Liuyi Jin}
\author[Address1]{Zhuo Zhao\corref{cor1}}
\cortext[cor1]{Corresponding author at: School of Mechanical Engineering, Xi’an Jiaotong University, Xi'an, Shaanxi, China.}

\address[Address1]{State Key Laboratory for Manufacturing System Engineering, Xi’an Jiaotong University,No.99 Yanxiang Road, Yanta District, 710054, Xi’an, Shaanxi, China}

\address[Address2]{International Joint Research Laboratory for Micro/Nano Manufacturing and Measurement Technologies, Xi’an Jiaotong University,No.99 Yanxiang Road, Yanta District, 710054, Xi’an, Shaanxi, China}

\begin{abstract}
Industrial visual inspection aims at detecting surface defects in products during the manufacturing process. Although existing anomaly detection models have shown great performance on many public benchmarks, their limited adjustability and ability to detect logical anomalies hinder their broader use in real-world settings. To this end, in this paper, we propose a novel component-aware anomaly detection framework (ComAD) which can simultaneously achieve adjustable and logical anomaly detection for industrial scenarios. Specifically, we propose to segment images into multiple components based on a lightweight and nearly training-free unsupervised semantic segmentation model. Then, we design an interpretable logical anomaly detection model through modeling the metrological features of each component and their relationships. Despite its simplicity, our framework achieves state-of-the-art performance on image-level logical anomaly detection. Meanwhile, segmenting a product image into multiple components provides a novel perspective for industrial visual inspection, demonstrating great potential in model customization, noise resistance, and anomaly classification. The code will be available at https://github.com/liutongkun/ComAD. 

\end{abstract}


\end{frontmatter}



\section{Introduction}
\label{}
Automated industrial visual inspection plays an important role in modern manufacturing quality control. Currently, it is popular to apply anomaly detection models in this field, because these models only need to collect normal samples for training and can theoretically detect any types of defects. In recent years, anomaly detection models have made significant progress \cite{roth2022towards,rudolph2023asymmetric, horwitz2022back, wang2023multimodal} in various benchmarks \cite{liu2023deep,xie2023iad}, both in terms of detection accuracy and efficiency. On the other hand, deploying these models in real-world scenarios is still challenging, which involves two important factors:

Firstly, existing models often lack sufficient adjustability to meet customized requirements. In practical applications, different manufacturers usually have different tolerance levels for different anomalies. For example, anomalies that appear on the product itself are typically more important than anomalies that appear in the image background. However, current models usually struggle to meet this simple requirement, since under the unsupervised setting,  it's difficult to inform the model about what type of anomaly is important through annotated information. Consequently, the anomaly score given by the model only reflects the degree to which the data deviates from the normal distribution established from a limited number of normal samples. This deviation may not necessarily correspond to human perception of `defects'. Indeed, `anomaly' is an objective concept from the perspective of data distribution, while `defect' involves more subjective human definitions. Although currently, human intervention is often undesirable for perception tasks in natural scenes, it may be significant for unsupervised industrial visual inspection, where the definition of defects is often subjective and variable. The anomaly detection model for real visual inspection should be more adjustable and explainable so that it can be easily transformed into a real `defect detection model' with the manufacturer's personal requirement.

\begin{figure}[t]
    \centering
		\includegraphics[width=0.9\columnwidth]{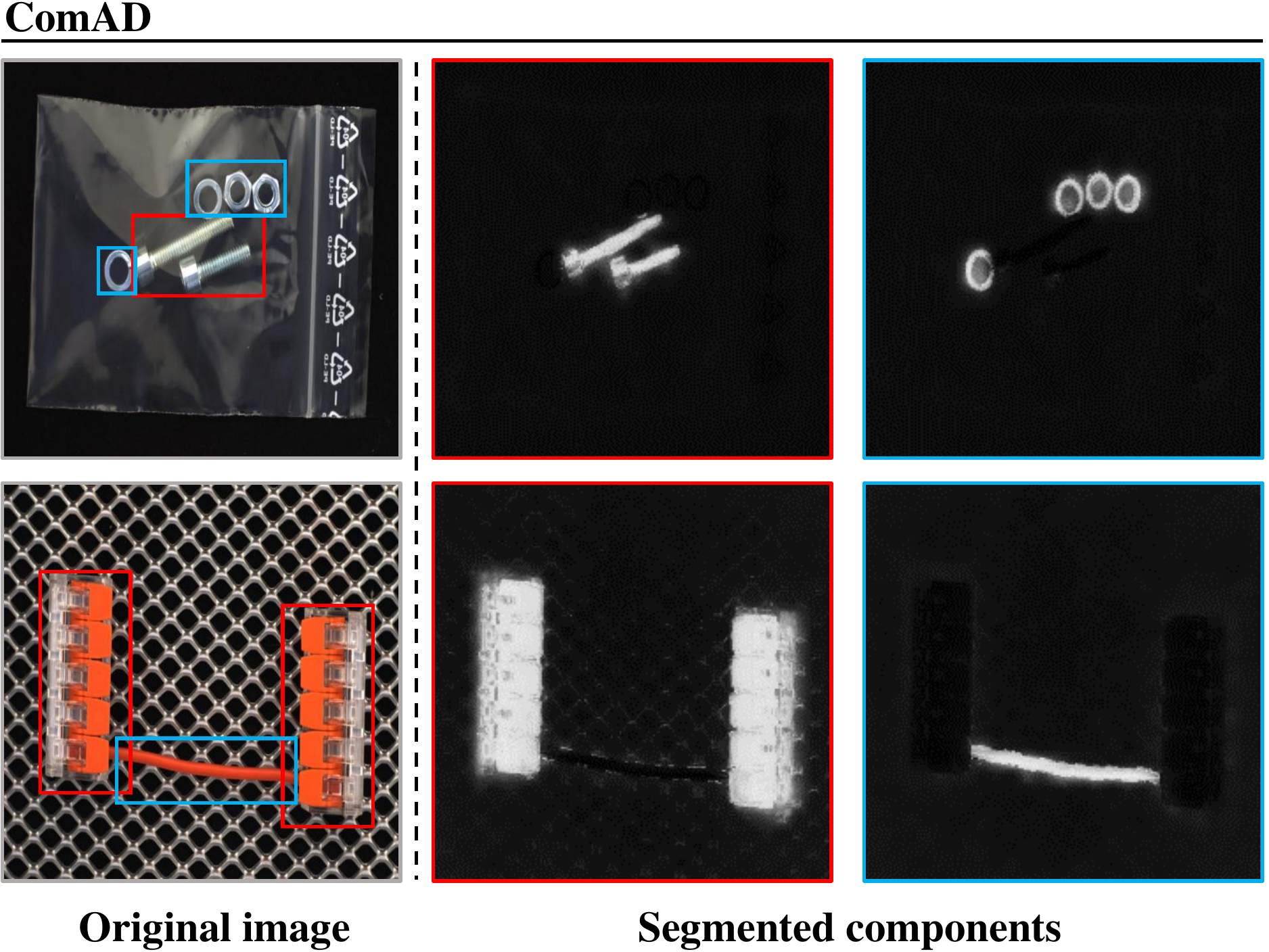}
	\caption{Examples from the MVTec LOCO AD dataset. Our method segments images into multiple components to achieve adjustable and logical anomaly detection.}
	\label{FIG:1}
\end{figure}

Secondly, existing models primarily address low-level structural anomalies, such as dents and scratches, while encountering challenges in detecting high-level semantic logical anomalies \cite{bergmann2022beyond}, such as missing components and incorrect component quantities, which require additional metrological features to measure or count the instance. However, achieving metrological analysis such as object counting can be very challenging under the unsupervised setting \cite{radford2021learning}. In practice, logical defects, such as component absence, may bring serious functional consequences, which are usually far more important than some scratches that only affect aesthetics. Nevertheless, the performance of current methods on logical anomaly detection benchmarks \cite{bergmann2022beyond, ishida2023sa} reveals considerable scope for enhancement.

There are several possible solutions for these two issues. For example, in terms of the adjustability of the model, anomaly clustering can be used \cite{sohn2023anomaly}, after which manufacturers could customize the model by changing the weights assigned to different anomaly categories. However, this approach cannot be used in cold-start scenarios, as it requires the model to have previously encountered abnormal samples. Moreover, due to the uncertainty inherent in anomalies, it may not guarantee that new anomaly types are accurately assigned weights. For logical anomaly detection, several specialized algorithms \cite{bergmann2022beyond, tzachor2023set, batzner2023efficientad} are proposed, which primarily rely on implicit deep global features to model long-range context. Though effective, these methods may also struggle to capture metrological information \cite{batzner2023efficientad} and they have weak interpretability and adjustability.

\begin{figure}[]

    \centering
		\includegraphics[width=0.9\columnwidth]{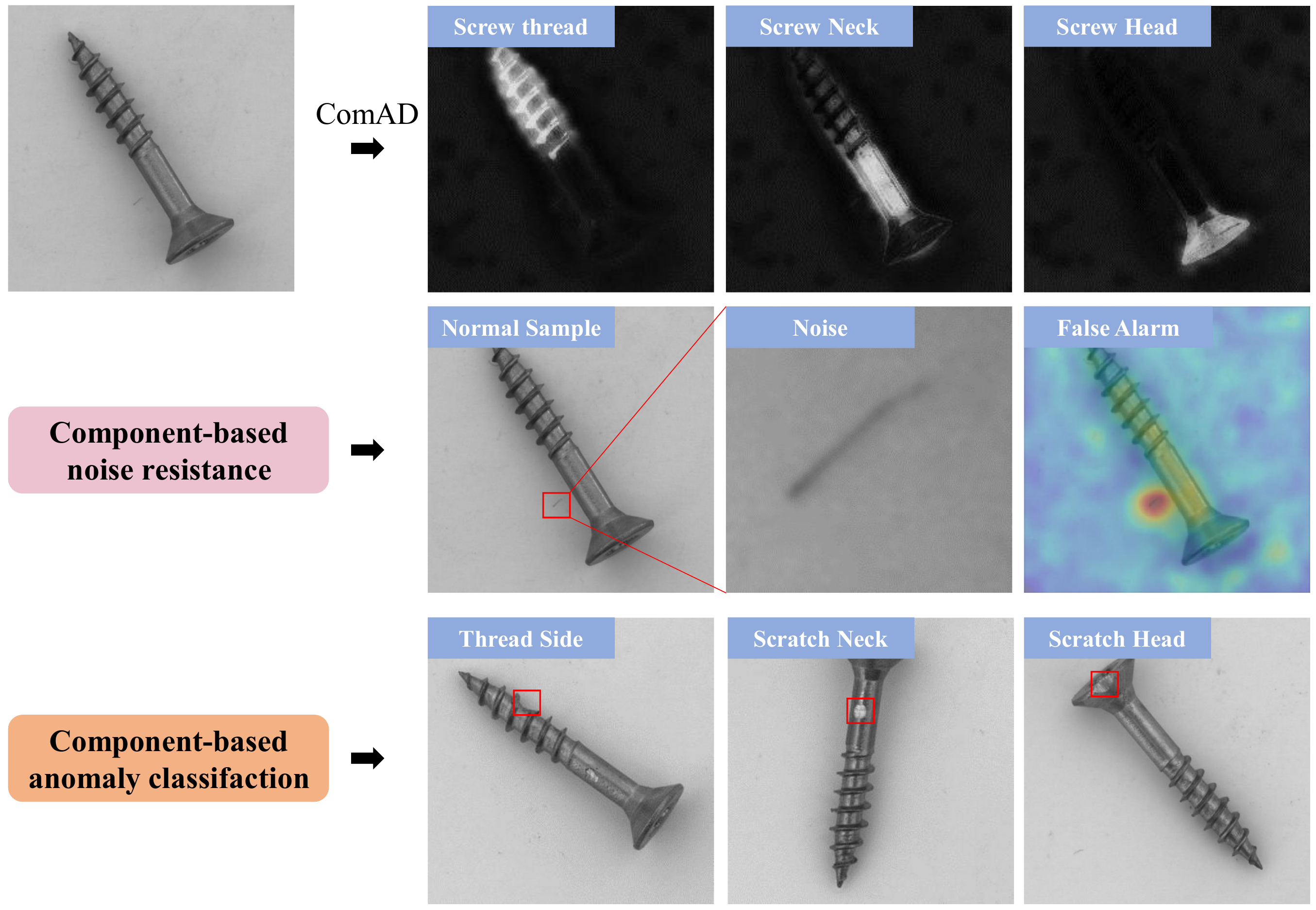}
	\caption{Examples from the MVTec AD screw dataset. By identifying the component to which the anomaly belongs, we can achieve anomaly classification and filter out background noise.}
	\label{FIG:2}
\end{figure}

Considering the aforementioned problems, in this paper, we propose a new component-aware anomaly detection framework for industrial visual inspection. Our method can serve as a plugin to simultaneously enhance the adjustability and the ability to detect logical anomalies of existing anomaly detection models. Specifically, we leverage DINO \cite{caron2021emerging} pre-trained features with post-processing algorithms to build a lightweight (1.9 MB) and almost training-free (less than one minute for training) unsupervised segmentation model, as shown in Fig. \ref{FIG:1}. Despite its simplicity, it enables effective and efficient image segmentation thanks to the high semantic DINO pre-trained features and the relatively homogeneous distribution of the industrial dataset. By integrating this component segmentation model with the existing anomaly detection model, we can identify anomalies in specific components, enabling more granular model adjustment based on the component's significance. Concretely, we can assign stricter thresholds to important components. Also, we can ignore anomalies that appear in irrelevant background areas, thereby improving the model's noise resistance. Besides, our segmentation model can help achieve anomaly classification, i.e., classify the anomalies based on the components they belong to rather than the anomalies themselves, thus better adapting to cold-start problems and new unknown anomalies. A typical example can be seen in Fig. \ref{FIG:2}, where the `Thread Side', `Scratch Neck', and `Scratch Head' in the third row are officially provided anomaly categories in the MVTec AD screw dataset \cite{bergmann2019mvtec}. For logical anomaly detection, our model mainly focuses on those metrological anomalies, e.g., the incorrect component quantities, which are typically difficult for existing methods. Based on our segmentation model, we design an algorithm to explicitly model the metrological features and therefore achieve effective and interpretable logical anomaly detection. Finally, our contributions are summarized as follows:

1. We provide a new perspective, i.e., segmenting the image into multiple components, to achieve adjustable and logical anomaly detection in industrial visual inspection. Experimentally, we have demonstrated its feasibility and advantages.

2. We propose a simple yet effective approach for unsupervised component segmentation and logical anomaly detection. 

3. Our logical anomaly detection model achieves state-of-the-art performance with interpretable results on the MVTec LOCO AD \cite{bergmann2022beyond} and CAD-SD \cite{ishida2023sa} datasets.

\section{Related Work}
\label{2}
\subsection{Industrial visual inspection}
\label{2.1}
Industrial visual inspection, also known as surface defect detection, has long been a widely researched topic. Early approaches \cite{xie2008review, kumar2008computer} mainly rely on handcrafted rule design, which offer strong interpretability and required few training samples. However, these methods are less generalizable and often encounter obstacles in complex scenarios. Later, researchers turned to supervised deep-learning models \cite{azizah2017deep,bovzivc2021mixed}. Although these models have improved upon the limitations of traditional handcrafted methods, they also introduce new issues, such as the need for a large number of defect samples for training and the difficulty of adapting to novel defect categories. As a result, in recent years, visual anomaly detection models are becoming increasingly popular, since they only require normal samples for training and can intrinsically generalize to outlier defects. Our method follows the basic setting of anomaly detection, where we only use normal samples for training without additional annotations.

\begin{figure*}[t]
    \centering
		\includegraphics[width=2\columnwidth]{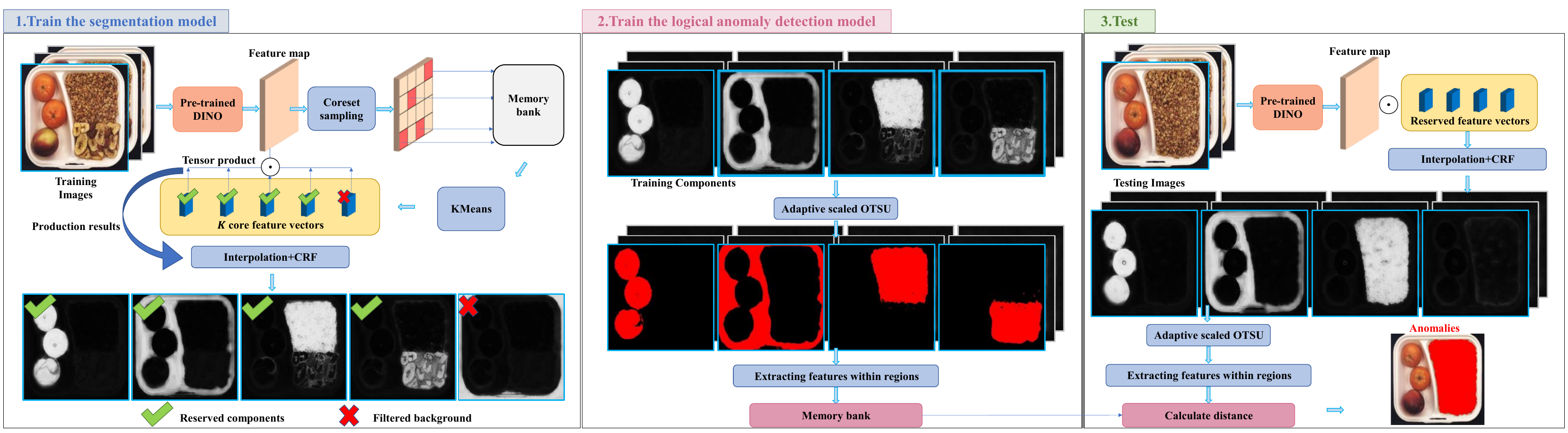}
	\caption{ComAD Overview. 1. We utilize KMeans clustering on the DINO pre-trained features to achieve component segmentations. 2. We extract regions from each segmentation map and further extract features within these regions to model the normal samples. 3. For test images, we compare their region features with those of training samples to achieve logical anomaly detection.}
	\label{FIG:3}
\end{figure*}

\subsection{Industrial visual anomaly detection}
\label{2.2}
Existing methods can be broadly classified into reconstruction-based and feature-based methods.

\textbf{Reconstruction-based methods} assume that models trained on normal samples can only reconstruct normal patterns, and will produce large reconstruction errors in abnormal regions. Typical models include autoencoders \cite{bergmann2018improving, zavrtanik2021reconstruction, zavrtanik2021draem, liu2022reconstruction, luo2023normal}, GANs \cite{schlegl2019f}, VAEs \cite{venkataramanan2020attention}, etc. Their advantage is that they have a certain level of interpretability, as people can analyze and adjust the model by observing the reconstructed images. On the other hand, the fuzzy generalization ability of reconstruction models often makes it difficult to find a good balance between reconstruction accuracy and distinguishability.  

\textbf{Feature-based methods} first project the image into a feature space through discriminative feature extractors. Due to the absence of negative samples for supervision, it's preferable to use ImageNet \cite{deng2009imagenet} pre-trained \cite{cohen2020sub} models or self-supervised \cite{li2021cutpaste} pre-trained model to extract features. The extracted features can be further modeled by gallery-based \cite{roth2022towards}, density-based \cite{rippel2021modeling}, flow-based \cite{rudolph2022fully}, and student-teacher-based models \cite{bergmann2020uninformed}, and the anomaly will be evaluated through either distance or density-based metrics. Overall, benefiting from the strong representation power of deep features, these methods typically outperform reconstruction-based algorithms. On the other hand, their interpretability is generally weaker.

Currently, both reconstruction-based and feature-based methods have achieved excellent performance on low-level structural anomalies, but face challenges in logical anomaly detection. Meanwhile, due to the lack of annotated information, compared to supervised models, it's more difficult to introduce prior knowledge to adjust anomaly detection models for customized requirements. Consequently, our framework does not focus on detecting structural anomalies but rather places more emphasis on model adjustability and logical anomaly detection capability.

\subsection{Unsupervised semantic segmentation and DINO pre-trained models}
\label{2.3}
Many unsupervised semantic segmentation approaches are designed for natural scenes. Some representative methods include IIC \cite{ji2019invariant}, PiCIE \cite{cho2021picie}, etc. Apart from convolutional neural network-based methods, Vision Transformer (ViT) \cite{dosovitskiy2020image} has shown advantages in long-range modeling. Specifically, Caron et al. \cite{caron2021emerging} (DINO) use self-distillation to train ViT in a self-supervised manner, and its self-attention shows meaningful semantics such as object boundaries. Hamilton et al. \cite{hamilton2022unsupervised} further distills the intermediate dense features of DINO to apply them to unsupervised semantic segmentation.

Our framework employs the unsupervised semantic segmentation model to segment industrial images into multiple components. Compared to the diverse images in natural scenes, the distribution of industrial images is relatively fixed, which reduces the requirement for the model's generalization ability. In particular, for the task of anomaly detection itself, a segmentation model with too strong a generalization ability may be harmful. Experimentally, we find that a simple clustering of the DINO pre-trained features can satisfy our requirements, while the above segmentation models may not bring additional benefits but even worse results. 

\section{Methods}
\label{3}
The proposed method is mainly designed for products with multiple components, without considering homogeneous textures, as the latter does not involve logical anomalies.
\subsection{Segmenting image into multiple components}
\label{3.1}
Given a training set of normal images $\mathcal{X}=\left \{ x_1,x_2,\dots x_i  \right \} $, we follow the strategy in \cite{hamilton2022unsupervised} to leverage the first block of the pre-trained DINO ViT-S/8 to calculate each image $x_i \in \mathbb{R}^{H \times W \times C}$'s intermediate feature, and denote it as $f_i \in \mathbb{R}^{I \times J \times D}$, where $(H, W)$ and $(I, J)$ represent spatial dimensions and $C, D$ represent channel dimensions. Then we perform corset sampling as \cite{roth2022towards} on each $f_i$ with a sampling ratio $r=0.01$ to remove redundant background features and reduce their storage cost, obtaining $f^{'}_{i} \in \mathbb{R}^{N\times D}$, where $N=\lfloor {r \times I \times J} \rfloor$ represents the number of reserved feature vectors. Remark that we do not perform dimensionality reduction on the feature vectors like some other methods \cite{roth2022towards, hamilton2022unsupervised} since we find it detrimental to final results. 

We concatenate all the $f^{'}_{i}$ in the training set $\mathcal{X}$ to build a memory bank $\mathcal{M} \in \mathbb{R}^{R \times D}$ where $R$ refers to the total number of all the reserved feature vectors. Then, we perform KMeans to cluster those feature vectors in $\mathcal{M}$ and therefore obtain $K$ clusters along with their center feature $f_{\mathrm{kmeans}} \in \mathbb{R}^{K\times D} $. For the original method, we employ $K=5$ in all the experiments.

Through computing the cosine similarity between $f_{\mathrm{kmeans}}$ and the original feature map $f_i$, we can obtain the preliminary segmentation map and then interpolate it to the size of the original image. Then we apply Fully connected Gaussian Conditional Random Fields (CRF) \cite{lafferty2001conditional} for post-processing to obtain our segmentation map $S \in \mathbb{R}^{H \times W \times K}$. For CRF, we use PyDenseCRF with 2 iterations. The parameters w.r.t the equation:
\begin{equation}
    w_{crf}(v_i,v_j)=a \ {\rm exp}(-\frac{\left |p_i-p_j  \right |^2 }{2\theta_{\alpha }^2}- \frac{\left |I_i-I_j  \right |^2 }{2\theta_{\beta  }^2})+b \ {\rm exp}(-\frac{\left |p_i-p_j  \right |^2 }{2\theta_{\gamma  }^2})
\end{equation}
are set as $a=4, b=3, \theta_{\alpha}=67, \theta_{\beta}=3, \theta_{\gamma} = 1$, which are the same as those in \cite{hamilton2022unsupervised}. 

Considering the uncertainty of unsupervised clustering, we do not apply `argmax' to $S$'s last dimension to assign a specific class to each pixel. Instead, we represent each sub segmentation map $s \in \mathbb{R}^{H \times W}$ separately for further refinement (discussed in the following section). Overall, the above process can be seen in Fig. \ref{FIG:3}.1. 
\begin{figure}
    \centering
		\includegraphics[width=\columnwidth]{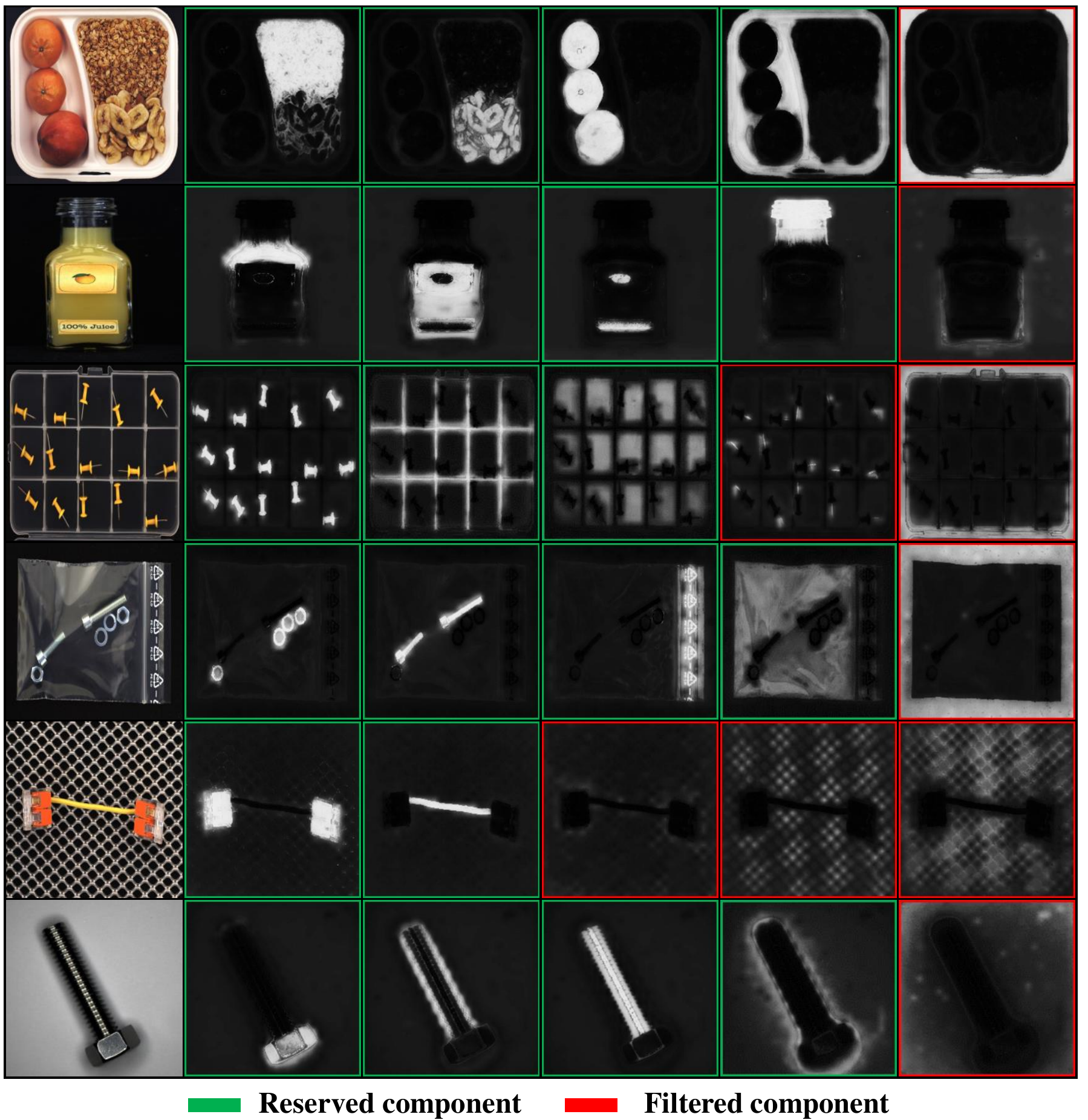}
	\caption{Examples of the segmented components under $K = 5$. }
	\label{FIG:4}
\end{figure}

\subsection{Component-based logical anomaly detection model}
\label{3.2}
Standardized manufactured products usually share uniform metrological features (the component size, quantities, etc.) within the same specification. Although these features cannot cover all types of anomalies, they are often the weak points of existing methods. For our logical anomaly detection model, we leverage the above segmentation map $S$ to capture the product's metrological features. Consequently, it can serve as an effective supplement to existing methods, allowing better performance on logical anomalies through ensemble detection. 

\subsubsection{Selecting the core components}
\label{3.2.1}
As the unsupervised segmentation task is inherently ambiguous, and the optimal value of $K$ is unknown when using KMeans, there may be some unexpected segmentation results in $S$. Therefore, we first choose to filter the noise and background components in $S$. To identify the noise component, we apply a mean filter with a size of $(11, 11)$ to each component map, and regard those maps whose maximum values are less than 0.5 (the value range of each map is [0, 1]) as noise components. To identify the background component, we apply OTSU \cite{otsu1979threshold} to convert each component map into regions, and adopt the strategy in \cite{wang2023cut}, where we suppose the background regions should occupy more than two corners of the image. Based on the above operations, our segmentation map undergoes a change from $S \in \mathbb{R}^{H \times W \times K}$ to $S \in \mathbb{R}^{H \times W \times K^{'}}$, where $K^{'}$ refers to the number of reserved categories. For the products involved in this paper, the aforementioned operations can be completed using only one training image. The retained component categories will be recorded and there is no need for redundant calculations during testing. Fig .\ref{FIG:4} illustrates some qualitative examples of the reserved and filtered components for several industrial products.

\begin{figure}
    \centering
		\includegraphics[width=\columnwidth]{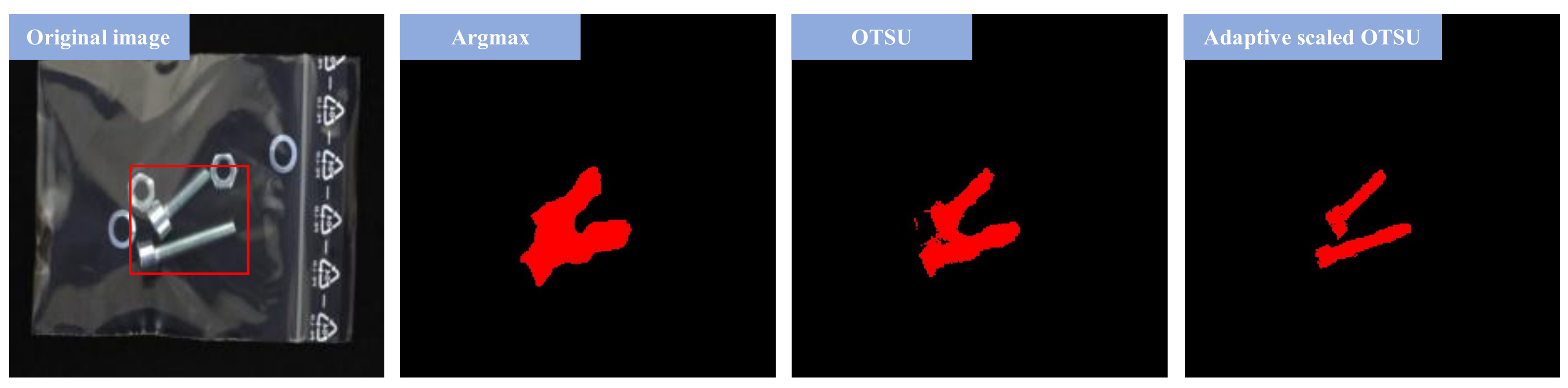}
	\caption{Qualitative comparisons of different region extraction algorithms.}
	\label{FIG:5}
\end{figure}

\subsubsection{Modeling the metrological features}
\label{3.2.2}
\textbf{Extracting component regions}. In order to obtain stable metrological features, we need to accurately extract the component region. In the field of image segmentation, a commonly used approach is to leverage the `argmax' function to assign specific labels to each pixel. For our segmentation model, we find it less effective. A better approach is to perform binary classification on each component map individually. Consequently, we leverage OTSU algorithm and make slight modifications to it. Specifically, OTSU can automatically determine the binary segmentation threshold by maximizing the inter-class variance between the foreground and background classes. Considering that our foreground may contain some noise, we apply a certain offset to the OTSU threshold to make the foreground criteria more stringent. To determine the offset, we employ a constraint that the optimal binary threshold should minimize the variance of the training set's region areas. Concretely, for each component's segmentation map $s \in \mathbb{R}^{H \times W}$, given the threshold calculated by the original OTSU  $\tau$, we scale it by each $c$ from $\{1, 1.1, 1.2, 1.3, 1.4\}$ respectively. Then we use each scaled $c\tau$ to binarize the image and calculate the foreground area variance of the entire training set. Finally, we choose the scaling factor $c^*$ which corresponds to the minimum variance. Also, $c^*$ will be recorded, so we do not need to recompute it during testing. We show some qualitative comparisons of `argmax', `OTSU', and our `Adaptive scaled OTSU' in Fig. \ref{FIG:5}.   

\textbf{Extracting component areas and colors}. After extracting the component region, we choose its area as our primary metrological feature. While simple, it can effectively indicate the component's states including the existence, quantities, etc. (We will also verify whether more complex deep features can capture these states in Sec \ref{5.1}.). However, we find that the DINO pre-trained features may group the objects with similar semantics but different colors into the same component category. For example, in the 1st row of Fig. \ref{FIG:4}, the `orange' and `peach' can not be distinguished by our segmentation model. Therefore, we further introduce the color features within each component region to refine the component discrimination.

For each component $k$, we denote its area and color features by $A^{k}$ and $Co^{k}$ respectively. Specifically, the area $A^{k}$ is calculated by the sum of the number of pixels in the region. To represent the color feature, we first convert the image from RGB to CIELAB space, which includes three components $L$, $a$, and $b$. We ignore $L$ to avoid the influence of lighting and only retain $b$ and $a$. For each pixel, we calculate $\frac{b}{a}$ and finally average that value over the entire region, thus obtaining $Co^{k}$. After obtaining the area and color features of each component, we concatenate them together to construct the global feature vector. Before this, we need to scale them to similar scales. Since a product may have multiple specifications, these features may not follow a Gaussian distribution, but instead a mixture of Gaussians with multiple centers. Therefore, we do not use standardization techniques such as z-score which require the data to follow a single Gaussian distribution. Meanwhile, to avoid sensitivity to noise, we do not apply Min-Max normalization. As a compromise, we normalize the feature by dividing it by its mean value in the training set, i.e., 
\begin{equation}
    {A^{'}}^{k}_{i}=\frac{A^{k}_i}{\sum_{n=0}^{N_{train}}A^{k}_n/N_{train}}
\end{equation}
\begin{equation}
    {Co^{'}}^{k}_{i}=\frac{Co^{k}_i}{\sum_{n=0}^{N_{train}}Co^{k}_n/N_{train}}
\end{equation}
where $i$ represents the $i$th image and $k$ represents its $k$th component. $N_{train}$ represents the number of training images. Therefore, for each original image $i$, we can obtain a global feature vector $G_i$, which is:
\begin{equation}
G_i=({{A^{'}}^1_i},{{Co^{'}}^1_i},{{A^{'}}^2_i},{{Co^{'}}^2_i},\dots, {{A^{'}}^k_i},{{Co^{'}}^k_i})
\end{equation}

We store all the $G_i$ in the training set. For the $j$th test image, we calculate its anomaly score $D_G$ using the average $l_2$ distance to its kNN (k-nearest neighbors), thus:
\begin{equation}
D_G= |G_{j} - \mathrm{kNN}_{G_{train}}(G_{j})|_2
\end{equation}
Based on $D_G$, we can obtain the image-level anomaly detection results. Meanwhile, by tracking back the specific components in $G_{j}$ that contribute to the $D_G$, we can obtain interpretable information about the component-level anomalies. 

\begin{figure}
    \centering
		\includegraphics[width=0.8\columnwidth]{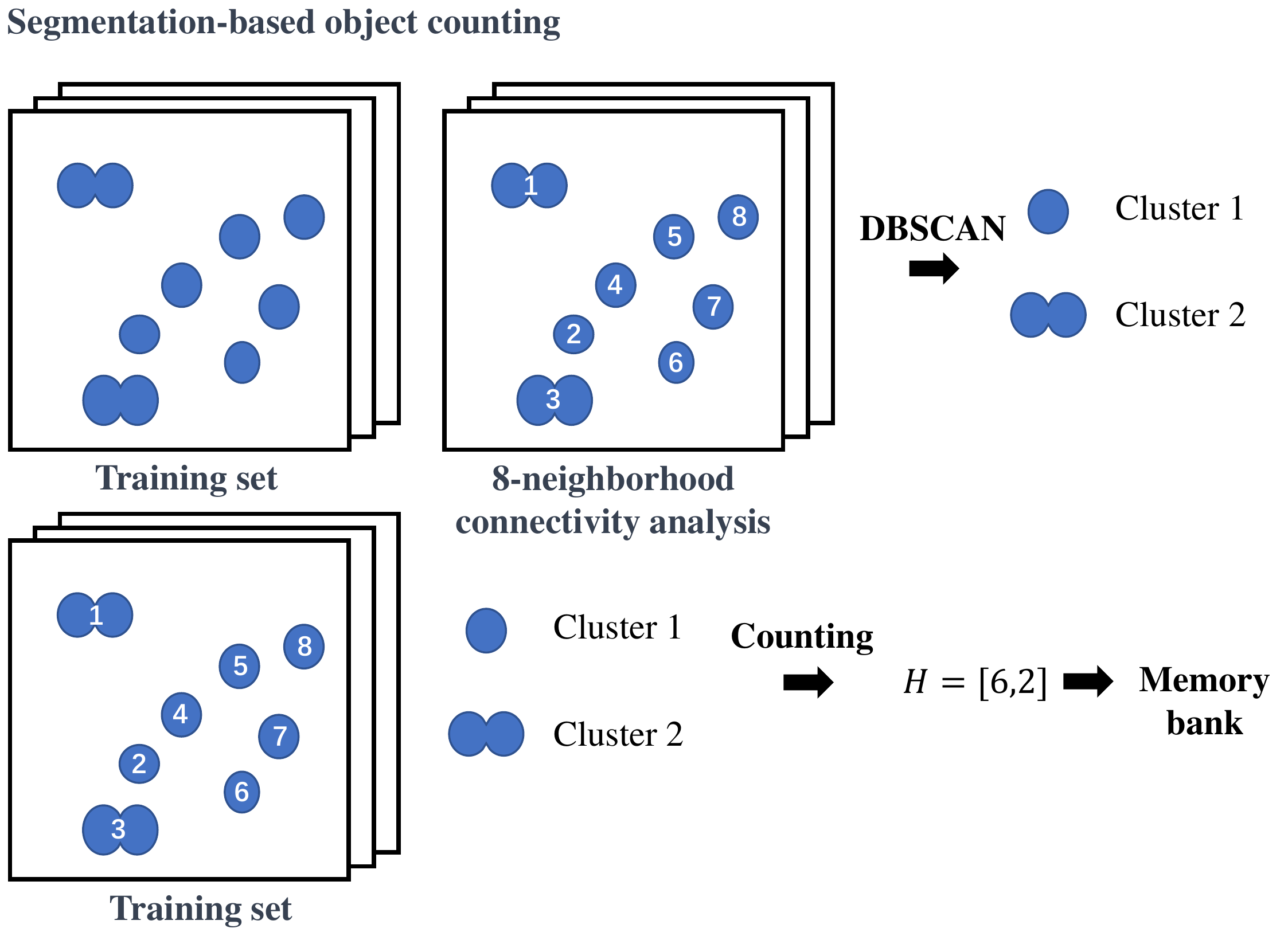}
	\caption{Illustration of object counting using region connectivity and DBSCAN clustering algorithms.}
	\label{FIG:6}
\end{figure}

\textbf{Object counting based on segmentation maps}. Through using area and color features, we can detect missing components or their incorrect quantities to some extent. However, this may not be suitable for the counting issue with a large number of instances. The main reason is that the area error of each instance, either from production error or segmentation error, accumulates, making the total area unable to reflect the accurate instance number. If these instances are not all morphologically adjacent, we can additionally leverage the connectivity of the regions to estimate the instance number, as can be seen in Fig. \ref{FIG:6}. Specifically, for the training set, we use the 8-connectivity criterion to divide the region into multiple separate regions and filter those tiny noises whose area is less than 0.1 \% of the image area. Then, we use DBSCAN \cite{ester1996density} algorithm to group these regions based on their area features. After that, each connected region can be assigned to a group based on which cluster it is closest to. Finally, for the $i$th image and its $k$th component, we count the number of instances in each group to obtain a histogram $H_i^k$, which approximately reflects the number of instances.

When the segmentation results are poor or there is significant instance variance, the histogram may have a high dimensionality, which affects the stability of detection. Therefore, we further regularize it based on the dimensionality $n_i^k$ of the histogram, i.e., the number of groups obtained by DBSCAN:
\begin{equation}
{H^{'}}_i^k = \frac{H_i^k}{n_i^k}
\end{equation}
Meanwhile, we do not concatenate the ${H^{'}}_i^k$ of each component into a global vector to avoid excessively high-dimensional feature vectors. Similarly, we store all the ${H^{'}}_i^k$ in the training set. For the $j$th test image, we calculate its anomaly score $D_H$ using the average $l_2$ distance to its kNN, thus:

\begin{equation}
D_H= \sum_{k=0}^{K^{'}}|{H^{'}}_j^k-\mathrm{kNN}_{H^k_{train}}({H^{'}}_j^k)|_2
\end{equation}

Overall, the complete anomaly score $D$ is represented as:
\begin{equation}
D= D_G + \alpha D_H
\end{equation}
where $\alpha$ is determined by the importance of object counting and in all experiments, we set $\alpha$ to 0.5. For the DBSCAN algorithm, we set its radius to be 10\% of the average area of the connected regions, and the minimum samples of 10. For all the kNN algorithms, we use 5-NN.

\section{Experiments}
\subsection{Datasets}
\label{4.1}
We primarily choose those datasets that involve logical anomalies to evaluate our logical anomaly detection model. Meanwhile, we evaluate the performance improvement brought by combining our method with other existing methods. Besides, we also select the benchmarks for structural anomalies so as to verify the overall performance. We use the area under the curve (AUC) of the receiver operating characteristics (ROC) as the evaluation metric.

\begin{figure}
    \centering
		\includegraphics[width=\columnwidth]{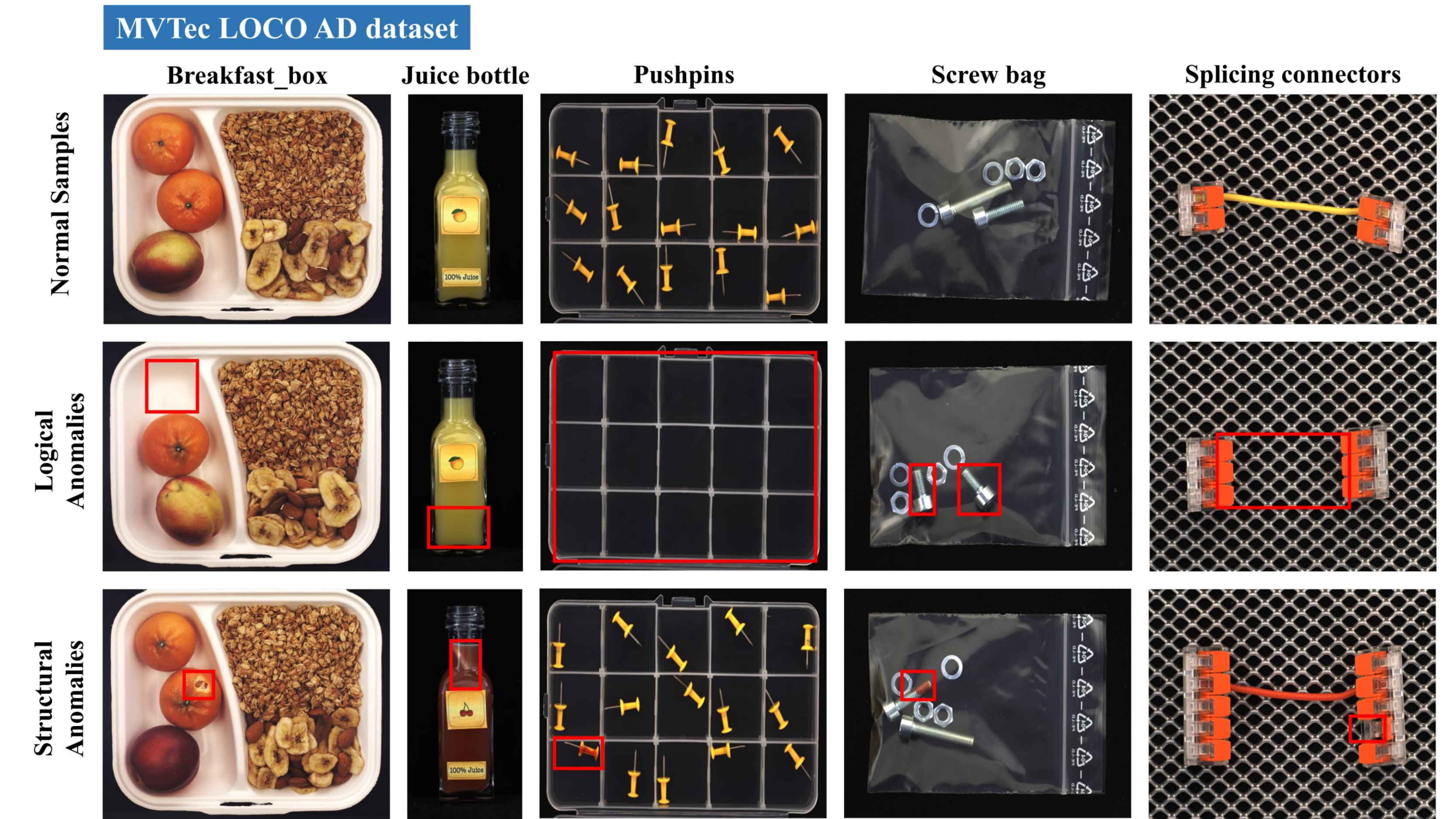}
	\caption{Example images of the MVTec LOCO AD dataset.}
	\label{FIG:7}
\end{figure}

\textbf{MVTec LOCO AD.} The MVTec LOCO AD dataset \cite{bergmann2022beyond} consists of five product categories and is mainly intended for logical anomaly detection. Specifically, logical anomalies differ from traditional structural anomalies in that the components themselves may appear to be in good condition (i.e., no visible scratches or dents), but at the product level, there may be errors in their quantities or assembly relationships. The dataset also involves structural anomalies. To clearly demonstrate the difference between logical and structural anomalies, we provide some examples in Fig. \ref{FIG:7}. The entire dataset involves 1772 normal images for training and 304 normal images for validation. For the test set, there are a total of 575 normal images, 432 structural anomaly images, and 561 logical anomaly images.

\begin{figure}
    \centering
		\includegraphics[width=0.6\columnwidth]{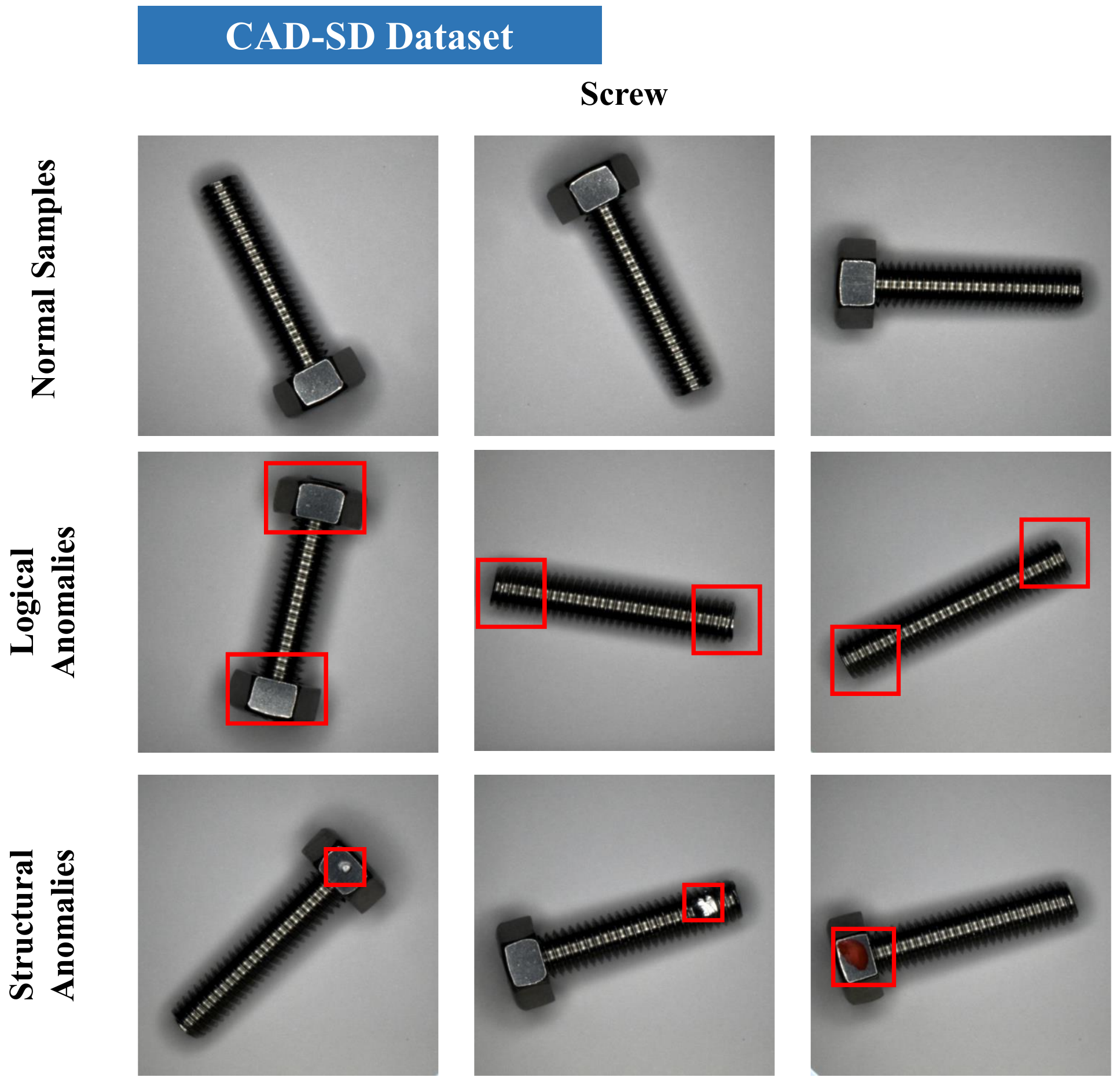}
	\caption{Example of images of the CAD-SD dataset.}
	\label{FIG:8}
\end{figure}

\textbf{Co-occurrence Anomaly Detection Screw Dataset (CAD-SD).} The CAD-SD \cite{ishida2023sa} involves a type of logical anomaly called `co-occurrence anomalies'. Specifically, this refers to the assembly relationship of screw rods and hex nuts. In the normal training set, a screw rod is constrained to be assembled with only one nut. Therefore, the violation of this constraint such as missing nut or double nut assembly will result in logical anomalies. Similarly, the dataset also includes structural anomalies of scratches and paint. The entire dataset involves 400 normal images for training. For the test set, there are a total of 210 normal images, 80 structural anomaly images, and 84 logical anomaly images. Some examples are shown in Fig. \ref{FIG:8}.

\begin{table*}[]
\centering
\caption{Quantitative comparisons of image-level detection results on the MVTec LOCO AD dataset. (AUROC\%)}
\label{Table1}
\scalebox{0.9}{
\begin{tabular}{c|ccc|cc|cc|cc|c|c}
\hline
\multirow{2}{*}{MVTec LOCO} & \multicolumn{1}{c|}{\multirow{2}{*}{Category}} & \multicolumn{2}{c|}{PatchCore} & \multicolumn{2}{c|}{RD4AD} & \multicolumn{2}{c|}{DRAEM} & \multicolumn{2}{c|}{AST} & GCAD & ComAD \\ \cline{3-12} 
 & \multicolumn{1}{c|}{} &  & +ComAD &  & +ComAD &  & +ComAD &  & +ComAD &  &  \\ \hline
\multirow{6}{*}{\begin{tabular}[c]{@{}c@{}}Logical\\ Anomalies\end{tabular}} & \multicolumn{1}{c|}{Breakfast Box} & 80.0 & 91.1 & 66.7 & 84.9 & 75.1 & 83.4 & 80.0 & 91.0 & - & \textbf{94.7} \\
 & \multicolumn{1}{c|}{Juice Bottle} & 92.3 & 95.0 & 93.6 & 95.5 & 97.8 & \textbf{98.8} & 91.6 & 93.9 & - & 90.9 \\
 & \multicolumn{1}{c|}{Pushpins} & 73.8 & \textbf{95.7} & 63.6 & 91.2 & 55.7 & 89.0 & 65.1 & 90.5 & - & 89.0 \\
 & \multicolumn{1}{c|}{Screw Bag} & 55.7 & 71.9 & 54.1 & 73.4 & 56.2 & 68.0 & 80.1 & \textbf{85.0} & - & 79.7 \\
 & \multicolumn{1}{c|}{Splicing Connectors} & 75.6 & \textbf{93.3} & 75.3 & 92.3 & 75.2 & 90.3 & 81.8 & 90.3 & - & 84.4 \\ \cline{2-12} 
 & Average.log & 75.5 & 89.4 & 70.7 & 87.5 & 72.0 & 85.9 & 79.7 & \textbf{90.1} & 86.0 & 87.7 \\ \hline
\multirow{6}{*}{\begin{tabular}[c]{@{}c@{}}Structural\\ Anomalies\end{tabular}} & \multicolumn{1}{c|}{Breakfast Box} & 75.2 & 81.6 & 60.3 & 69.1 & 85.4 & \textbf{86.2} & 79.9 & 80.6 & - & 70.0 \\
 & \multicolumn{1}{c|}{Juice Bottle} & 97.8 & \textbf{98.2} & 95.2 & 97.8 & 90.8 & 93.2 & 95.5 & 96.6 & - & 80.5 \\
 & \multicolumn{1}{c|}{Pushpins} & 81.9 & 91.1 & 84.8 & 91.9 & 81.5 & 89.9 & 77.8 & 93.1 & - & \textbf{93.8} \\
 & \multicolumn{1}{c|}{Screw Bag} & 88.6 & 88.5 & 89.2 & 89.2 & 85.0 & 84.8 & \textbf{95.9} & 87.4 & - & 65.0 \\
 & \multicolumn{1}{c|}{Splicing Connectors} & 94.9 & 94.9 & \textbf{95.9} & \textbf{95.9 }& 95.5 & 95.2 & 89.4 & 89.3 & - & 63.8 \\ \cline{2-12} 
 & \multicolumn{1}{c|}{Average.str} & 87.7 & \textbf{90.9} & 85.1 & 88.8 & 87.6 & 89.9 & 87.7 & 89.4 & 80.6 & 74.6 \\ \cline{2-12} 
 & \multicolumn{1}{c|}{Average} & 81.6 & \textbf{90.1} & 77.9 & 88.2 & 79.8 & 87.9 & 83.7 & 89.8 & 83.3 & 81.2 \\ \hline
\end{tabular}}
\end{table*}

\begin{table*}[]
\caption{Quantitative comparisons of image-level detection results on the CAD-SD dataset. (AUROC\%)}
\centering
\label{Table2}
\scalebox{0.9}{
\begin{tabular}{c|c|cc|cc|cc|cc|cc|c}
\hline
\multirow{2}{*}{CAD-SD} & \multirow{2}{*}{Category} & \multicolumn{2}{c|}{PatchCore} & \multicolumn{2}{c|}{RD4AD} & \multicolumn{2}{c|}{DRAEM} & \multicolumn{2}{c|}{AST} & \multicolumn{2}{c|}{SA-PatchCore} & ComAD \\ \cline{3-13} 
 &  &  & +ComAD &  & +ComAD &  & +ComAD &  & +ComAD &  & +ComAD &  \\ \hline
\begin{tabular}[c]{@{}c@{}}Logical\\ Anomalies\end{tabular} & Screw & 64.6 & \textbf{100.0} & 48.7 & \textbf{100.0} & 46.9 & \textbf{100.0} & 81.7 & \textbf{100.0} & 98.8 & \textbf{100.0} & \textbf{100.0} \\ \hline
\begin{tabular}[c]{@{}c@{}}Structural\\ Anomalies\end{tabular} & Screw & \textbf{100.0} & 99.7 & 99.7 & 99.7 & 97.8 & 98.1 & 97.3 & 97.4 & 95.9 & 97.2 & 81.3 \\ \hline
 & Average & 82.3 & \textbf{99.9} & 74.2 & \textbf{99.9} & 72.4 & 99.1 & 89.5 & 98.7 & 97.4 & 98.6 & 90.7 \\ \hline
\end{tabular}}
\end{table*}

\begin{table*}[]
\caption{Quantitative comparison of image-level detection results on multiple benchmarks. (AUROC\%)}
\centering
\label{Table3}
\begin{tabular}{c|cc|cc|cc|cc}
\hline
\multirow{2}{*}{Datasets} & \multicolumn{2}{c|}{PatchCore} & \multicolumn{2}{c|}{RD4AD} & \multicolumn{2}{c|}{DRAEM} & \multicolumn{2}{c}{AST} \\ \cline{2-9} 
 &  & +ComAD &  & +ComAD &  & +ComAD &  & +ComAD \\ \hline
MVTec LOCO Logical & 75.5 & 89.4 & 70.7 & 87.3 & 72.0 & 86.0 & 79.7 & \textbf{90.2} \\
MVTec LOCO Structural & 87.7 & \textbf{90.9} & 85.1 & 89.0 & 87.6 & 89.9 & 87.7 & 89.4 \\
CAD-SD & 82.3 & \textbf{99.9} & 74.2 & \textbf{99.9} & 72.4 & 99.3 & 89.5 & 98.9 \\
MVTec AD Object & \textbf{99.2} & 97.7 & 98.0 & 97.4 & 97.5 & 96.0 & 98.5 & 97.0 \\\hline
Average & 86.2 & \textbf{94.5} & 82.0 & 93.4 & 82.4 & 92.8 & 88.9 & 93.9 \\ \hline
\end{tabular}
\end{table*}

\textbf{MVTec AD.} The MVTec AD \cite{bergmann2019mvtec} dataset consists of 15 different industrial products, including 10 object and 5 texture categories. Their anomaly detection task primarily focuses on traditional structural defects. The entire dataset involves 3629 normal images for training. For the test set, there are a total of 467 normal images and 1258 abnormal images. The proposed method is solely evaluated on the object categories without considering the homogeneous texture categories.   

\subsection{Implementation details}
\label{4.2}
We implement all the experiments with an NVIDIA GeForce GTX 3090TI and I7-12700 in Pytorch. For our ComAD, We resize all the images into $224 \times 224$ and each image takes approximately 40 ms for the component segmentation and 0.007 ms for anomaly detection. We five-run all the experiments and take the average value as the result. 

\subsection{Results}
\label{4.3}
We choose several state-of-the-art industrial anomaly detection models as our baselines, including reconstruction-based DRAEM \cite{zavrtanik2021draem}\footnote{https://github.com/VitjanZ/DRAEM.}, and feature-based AST \cite{rudolph2023asymmetric}\footnote{https://github.com/marco-rudolph/AST}, PatchCore \cite{roth2022towards}\footnote{https://github.com/amazon-science/patchcore-inspection}, and RD4AD \cite{deng2022anomaly}\footnote{https://github.com/hq-deng/RD4AD}. We also compare our method with the specific logical anomaly detection model GCAD \cite{bergmann2022beyond} and SA-PatchCore \cite{ishida2023sa}\footnote{https://github.com/IshidaKengo/SA-PatchCore}. For the chosen baselines \cite{zavrtanik2021draem, rudolph2023asymmetric, roth2022towards, deng2022anomaly, ishida2023sa}, we implement them with their official codes. For Patchcore, we use the WideResNet-50 \cite{he2016deep} as the backbone. For GCAD, as there is no official code, we directly report the performance of their paper. To combine our ComAD with other baselines, we simply add their anomaly scores. The results on the MVTec LOCO AD and CAD-SD datasets are shown in Table. \ref{Table1} and Table. \ref{Table2} respectively.

\begin{figure}
    \centering
		\includegraphics[width=0.9\columnwidth]{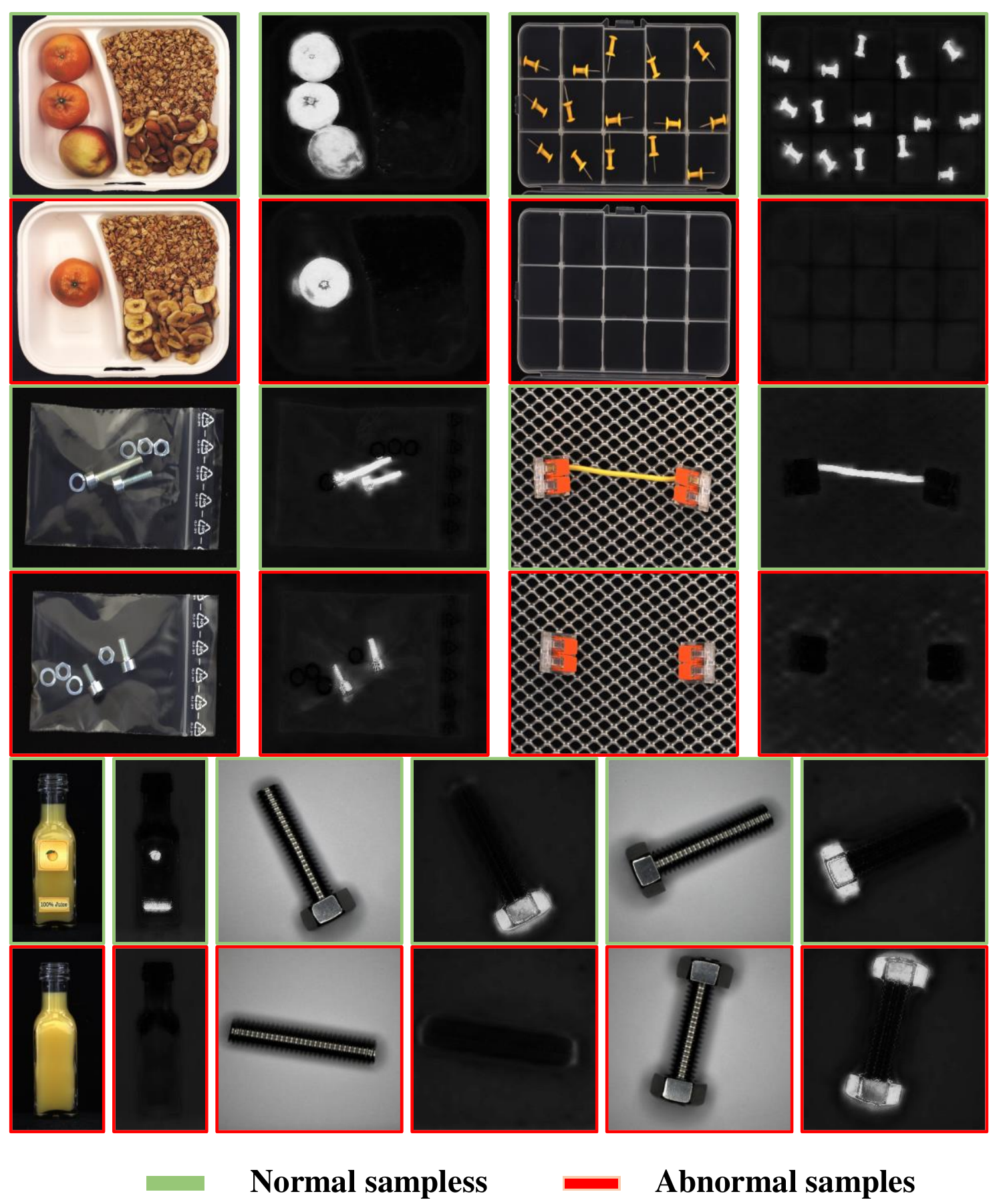}
	\caption{Qualitative comparisons of segmented components in normal and logical abnormal samples.}
	\label{FIG:9}
\end{figure}

Based on our results, it can be observed that the existing methods generally perform poorly in logical anomaly detection except for the `juice bottle' category, whose normal samples share a relatively small intra-class variance. When the normal category has a larger intra-class variance, such as the rotated screw in the CAD-SD datasets, even if it looks very simple, existing methods struggle with its logical anomalies. For the multiple component products, the variance will be even larger, e.g., if each component has a certain degree of rotational variation, their combinations will generate countless image patterns. In such scenarios, our method can reduce the image complexity by decoupling the product into individual components. Meanwhile, those metrological features such as the area of the component are less susceptible to the component's pose transformations. Therefore, when ComAD is used alone, it achieves state-of-the-art performance in logical anomaly detection tasks. We provide qualitative comparisons between normal and logically abnormal samples' segmentation maps in Fig. \ref{FIG:9}, where we can observe significant differences. On the other hand, since our model mainly considers those metrological features, it can not detect other types of logical anomalies. For example, in the `splicing connectors' category, many logical anomalies manifest as structural variations, such as the breakage or misalignment of the wire. In this case, combining our model with existing baselines for ensemble detection can achieve better performance. Besides, the performance is still unsatisfactory in the `screw bag' category, even though its logical anomalies are all caused by the incorrect number or size of components. We attribute its reasons to 1. the area differences between each component are relatively large, either due to occlusion or due to projection transformation during imaging; 2. for small annular objects such as washers and nuts, our segmentation results are ambiguous where sometimes only the ring is considered, while other times the entire object is included; 3. our segmentation model cannot make a more fine-grained distinction between the components `washer' and `nut'. In this situation, it may be preferable to use instance-level perception algorithms.   
   
In terms of structural anomalies, our method is unable to detect fine-grained structural differences, but can to some extent detect larger structural defects. For example, our method achieves the best structural anomaly detection results in the `pushpins' category, where many structural defects are manifested as large color differences or incomplete components.

For ensemble detection, our method significantly improves the performance of existing models on the MVTec LOCO AD and CAD-SD datasets. To verify its stability, we further conduct experiments on the MVTec AD dataset, which mainly consists of fine-grained structural anomalies. As shown in Table. \ref{Table3}, we observe a performance decrease on the MVTec AD Object dataset. This is because the object categories in the MVTec AD dataset are mainly composed of single-component products, which may lead to an over-segmentation with the original segmentation number $K$. We will discuss this in our ablation studies (Sec \ref{5.2}, Table. \ref{Table5}). In general, integrating our method with existing models leads to significant performance improvements. Meanwhile, it offers more model adjustability, as previously mentioned in Fig. \ref{2}.    

\section{Discussion}
\subsection{The leverage of the pre-trained features}
\label{5.1}
Using ImageNet supervised pre-trained features is a common approach in industrial anomaly detection. In this paper, however, we utilize self-supervised DINO pre-trained features to segment images into multiple components. From the perspective of feature reuse, we can consider the following questions: 1. Can ImageNet supervised features be clustered for unsupervised segmentation? 2. Can DINO pre-trained features be used for structural anomaly detection? Besides, for logical anomaly detection, we leverage the explicit area features to capture the component's metrological information. Therefore, we will also explore whether this information can be implicitly captured by the deep pre-trained features.

\begin{figure}
    \centering
		\includegraphics[width=0.9\columnwidth]{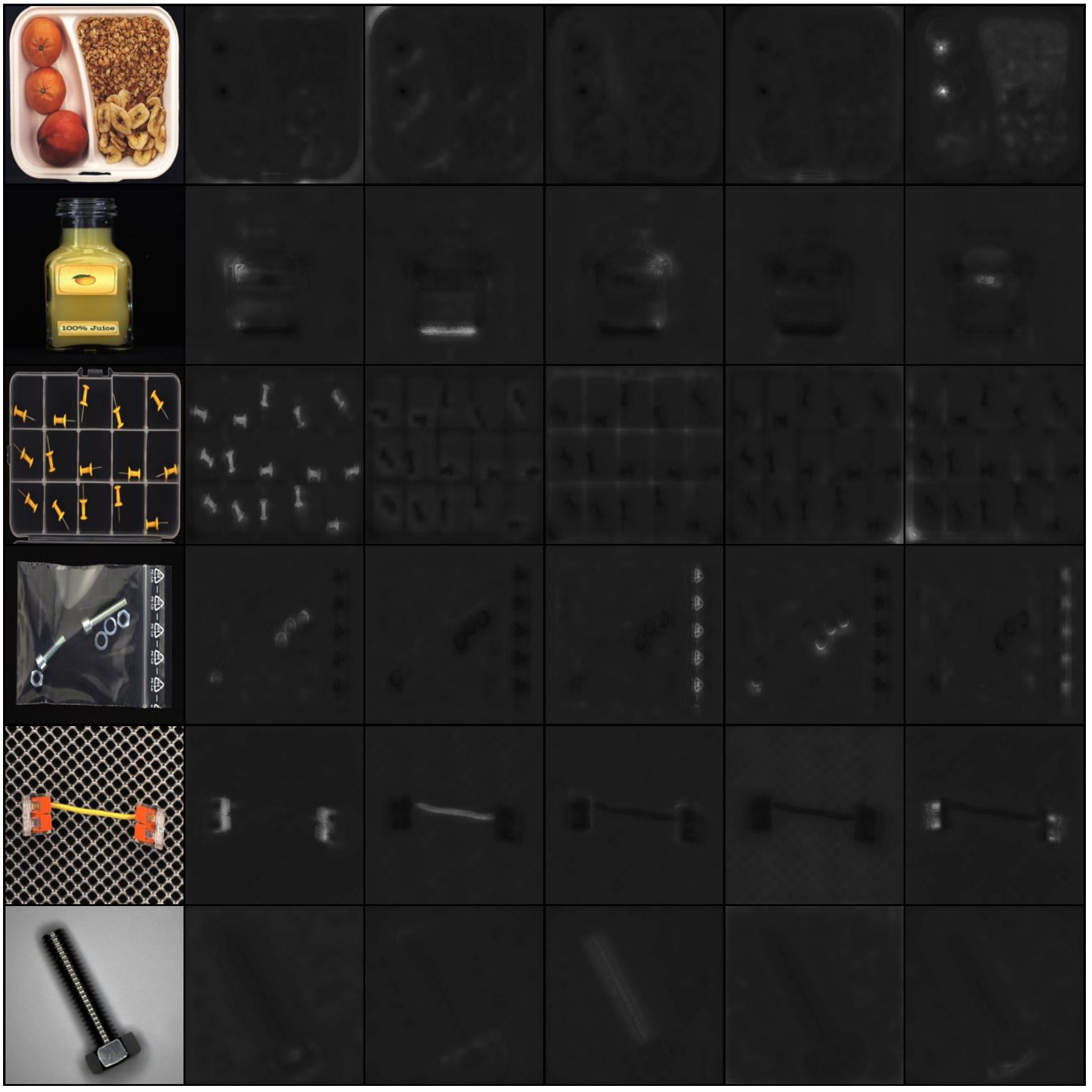}
	\caption{Segmentation results based on supervised pre-trained feature clustering.}
	\label{FIG:10}
\end{figure}

\textbf{Can ImageNet supervised features be clustered for unsupervised segmentation?} We fix other settings and only replace the pre-trained backbone in ComAD with an ImageNet supervised pre-trained WideResNet-50. Specifically, we use the features from `layer 2'. The segmentation results are shown in Fig. \ref{FIG:10}. As compared to Fig. \ref{FIG:4}, we observe poor results with lower contrast and meaningless components in most cases. Among them, the results of the `splicing connectors' and `pushpins' categories are marginally acceptable. We also attempt to use deeper layer features including `layer3' and `layer4', but experimentally, they do not bring any improvement.

\textbf{Can DINO pre-trained features be used for structural anomaly detection?} We leverage PatchCore and replace its extracted features with the DINO pre-trained intermediate features from its first block. We denote it as $\mathrm{Patchcore}_{\mathrm{DINO}}$ and do experiments on the MVTec AD dataset. As shown in Table. \ref{Table4}, we find that the DINO pre-trained features are not discriminative for low-level structural differences. Similarly, selecting deeper transformer blocks does not lead to any improvement.  

\begin{table}[]
\caption{Quantitative comparisons of image-level detection results for Patchcore with different pre-trained features on the MVTec AD dataset. (AUROC\%)}
\label{Table4}
\centering
\scalebox{0.9}{
\begin{tabular}{c|cc}
\hline
MVTec AD dataset & PatchCore & $\mathrm{Patchcore}_{\mathrm{DINO}}$ \\ \hline
Carpet           & \textbf{98.4}      & 57.7          \\
Grid             & \textbf{98.0}      & 74.0          \\
Leather          & \textbf{100.0}     & 55.3          \\
Tile             & \textbf{99.1}      & 80.2          \\
Wood             & \textbf{99.0}      & 64.3          \\
Bottle           & \textbf{100.0}     & 99.7          \\
Cable            & \textbf{99.8}      & 87.5          \\
Capsule          & \textbf{97.9}      & 84.7          \\
Hazelnut         & \textbf{100.0}     & 62.0          \\
Metal Nut        & \textbf{99.9}      & 85.3          \\
Pill             & \textbf{96.6}      & 46.2          \\
Screw            & \textbf{98.7}      & 58.0          \\
Toothbrush       & \textbf{99.7}      & 89.4          \\
Transistor       & \textbf{100.0}     & 83.8          \\
Zipper           & \textbf{99.5}      & 95.0          \\ \hline
Average          & \textbf{99.1}      & 74.9          \\ \hline
\end{tabular}}
\end{table}

\begin{figure}
    \centering
		\includegraphics[width=0.9\columnwidth]{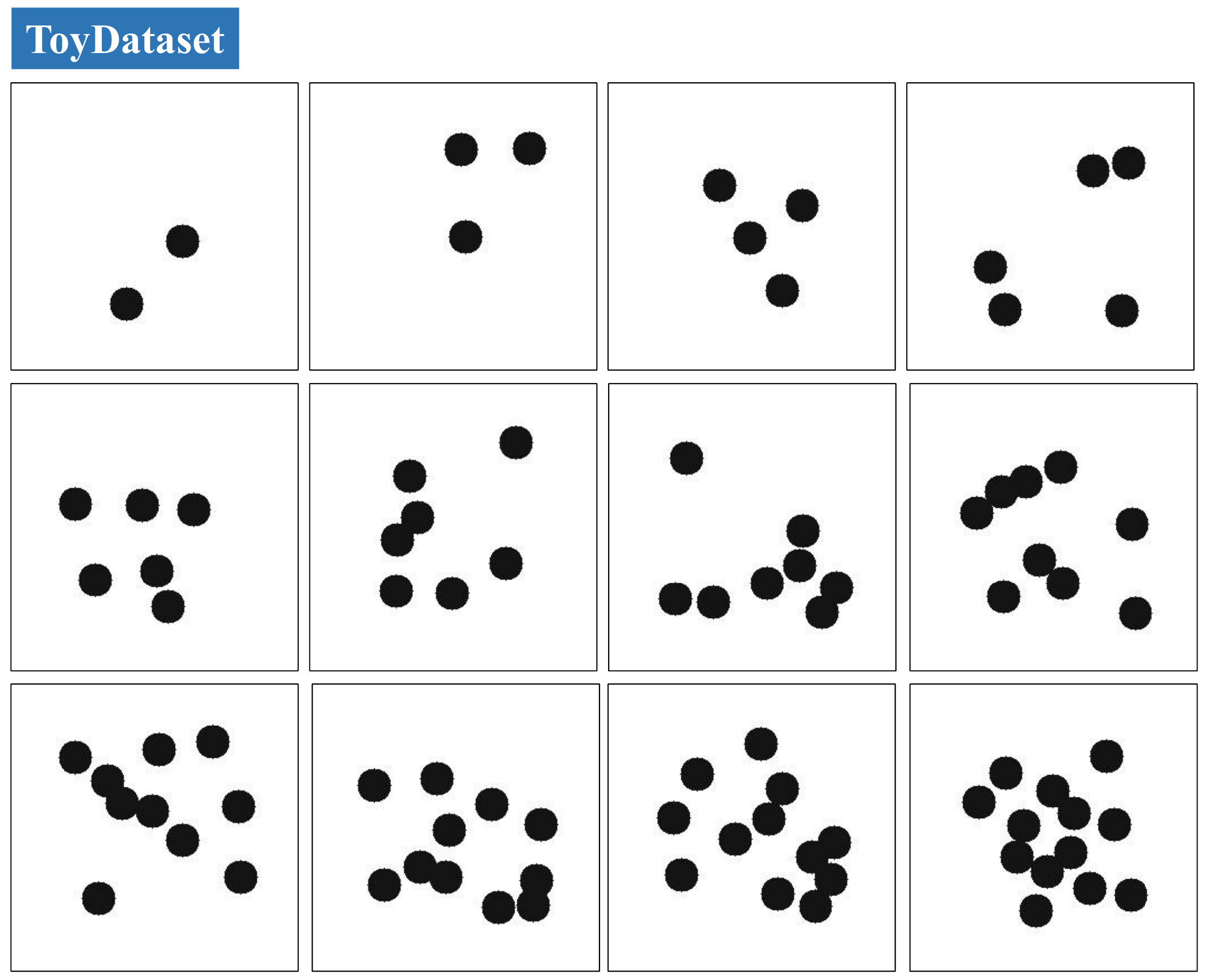}
	\caption{A toy dataset for validating whether deep pre-trained features can capture metrological information. 
There are a total of 12 categories, and within each category, the number of circles in the images is consistent.}
	\label{FIG:11}
\end{figure}

\textbf{Can deep pre-trained features capture the metrological information?} We create a toy dataset that is made of the same circle element but with different quantities, as shown in Fig. \ref{FIG:11}. Specifically, the dataset is divided into 12 categories based on the number of circles (from 2 to 13) in each image. Each category contains 100 images with the dimension of $256 \times 256 \times 3$. The pixel value of all the circles is [20, 20, 20], and their radius is 15. For each image, the positions of the circles are randomized. The circles may be adjacent to each other but will not overlap excessively.      

\begin{figure*}
    \centering
		\includegraphics[width=2\columnwidth]{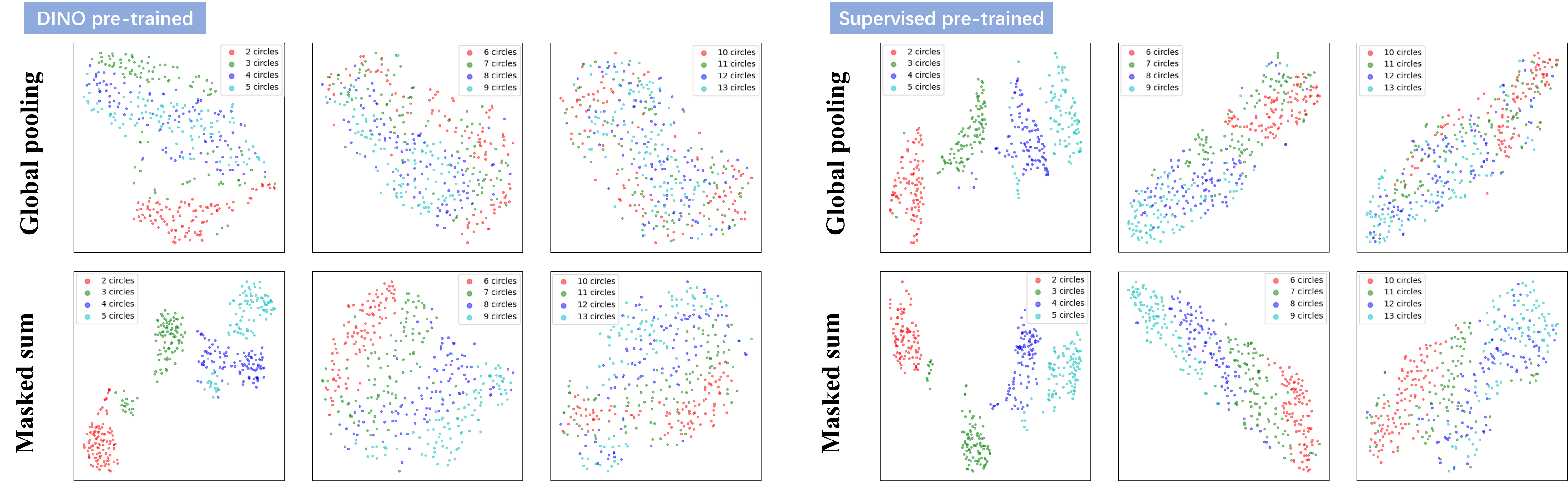}
	\caption{
Visualization of the image distributions in our toy dataset under different pre-trained features and feature extraction strategies.}
	\label{FIG:12}
\end{figure*}

We attempt two strategies to model the image. 1. We simply employ global average pooling to obtain a global feature vector of each image. 2. Assuming that the regions of the circles are already known, we only sum the feature vectors within the circle regions (we denote it as `Masked sum'). For the deep pre-trained features, we leverage the WideResNet-50 `layer2' and the intermediate features from the first block of DINO as above. Finally, we visualize the distribution of each category using t-SNE \cite{van2008visualizing}, as shown in Fig. \ref{FIG:12}.

From the results, we can observe that the deep pre-trained features can only identify the metrological difference when the number of circles is small. As the number of circles increases, they become ineffective. Comparatively, `Masked sum' performs relatively better than global pooling, and supervised features perform relatively better than DINO's features. However, they do not bring substantial changes. Meanwhile, we also attempt to select features from different hierarchical levels, but the issue remains unresolved. 

In conclusion, our experiments show that while DINO pre-trained features can capture high-level semantics for component-level perception, it may ignore fine-grained texture differences and is therefore not well suited for low-level structural anomaly detection tasks. In contrast, ImageNet supervised pre-trained features exhibit nearly opposite characteristics. In addition, both of them are difficult to directly apply to metrological statistical problems. Besides, we find that the features from different blocks in DINO perform similarly in our tasks, therefore, for simplicity, we only select the features in the first block.

\begin{table*}[]
\caption{Quantitative comparison of image-level detection results with different $K$ values on multiple benchmarks. (AUROC\%)}
\centering
\label{Table5}
\begin{tabular}{c|cc|cc|cc|c}
\hline
\multirow{2}{*}{Datasets} & \multicolumn{2}{c|}{$K=3$} & \multicolumn{2}{c|}{$K=4$} & \multicolumn{2}{c|}{$K=5$} & \multirow{2}{*}{Patchcore} \\
                          & ComAD    & +PatchCore    & ComAD    & +PatchCore    & ComAD    & +PatchCore    &                            \\ \hline
MVTec LOCO Logical        & 84.1     & 85.9          & 87.7     & 89.1          & 87.7     & \textbf{89.4}          & 75.5                       \\
MVTec LOCO Structural     & 71.0     & 89.6          & 73.1     & \textbf{90.9}          & 74.6     & \textbf{90.9}          & 87.7                       \\
CAD-SD Logical            & 99.5     & 95.5          & \textbf{100}      & \textbf{100}           & \textbf{100}      & \textbf{100}           & 64.6                       \\
CAD-SD Structural         & 53.6     & \textbf{100}           & 70.8     & \textbf{100}           & 81.3     & 99.7          & \textbf{100}                        \\
MVTec AD Object           & 67.9     & \textbf{99.2}          & 72.7     & 98.3          & 70.2     & 97.7          & \textbf{99.2}                       \\ \hline
Average                   & 75.2     & 94.0          & 80.9     & \textbf{95.7}          & 82.8     & 95.5          & 85.4                       \\ \hline
\end{tabular}
\end{table*}

\begin{table}[]
\caption{Quantitative comparisons of image-level detection results with (w/) and without (w/o) CRF on the logical anomaly detection benchmarks. The results are represented as (logical anomaly detection AUROC \%, structural anomaly detection AUROC \%) }
\centering
\label{Table6}
\scalebox{0.9}{
\begin{tabular}{c|cc}
\hline
Category            & w/ CRF    & w/o CRF      \\ \hline
Breakfast Box       & (\textbf{94.5}, \textbf{70.0}) & (92.7, 63.3) \\
Juice Bottle        & (\textbf{90.8}, \textbf{80.5}) & (89.5, 78.0) \\
Pushpins            & (\textbf{89.0}, \textbf{93.8}) & (85.2, 88.5) \\
Screw Bag           & (79.9, 65.0) & (\textbf{84.5}, \textbf{69.5}) \\
Splicing Connectors & (\textbf{84.3}, \textbf{63.8}) & (82.8, 59.4) \\ \hline
Average.MVTec LOCO & (\textbf{87.7}, \textbf{74.6}) & (86.9, 71.7) \\ \hline
Screw(CAD-SD)       & (\textbf{100}, 81.3)  & (\textbf{100}, \textbf{85.9})  \\ \hline
\end{tabular}}
\end{table}

\begin{table}[]
\caption{Quantitative comparisons of image-level detection results with different region extraction methods (`ASO' refers to `Adaptive scaled OTSU' ) on the logical anomaly detection benchmarks. The results are represented as (logical anomaly detection AUROC \%, structural anomaly detection AUROC \%) }
\centering
\label{Table7}
\scalebox{0.8}{
\begin{tabular}{c|ccc}
\hline
Category            & argmax       & OTSU         & ASO  \\ \hline
Breakfast Box       & (\textbf{94.5}, 66.2) & (\textbf{94.5}, \textbf{70.0}) & (\textbf{94.5}, \textbf{70.0}) \\
Juice Bottle        & (84.2, 75.0) & (\textbf{\textbf{90.8}}, \textbf{80.5}) & (\textbf{\textbf{90.8}}, \textbf{80.5}) \\
Pushpins            & (79.4, 72.5) & (87.0, 93.5) & (\textbf{89.0}, \textbf{93.8}) \\
Screw Bag           & (\textbf{84.0}, \textbf{74.6}) & (77.5, 64.6) & (79.9, 65.0) \\
Splicing Connectors & (84.2, 59.7) & (\textbf{84.3}, \textbf{63.8}) & (\textbf{84.3}, \textbf{63.8}) \\ \hline
Average.MVTec LOCO & (85.3, 69.6) & (86.8, 74.5) & (\textbf{87.7}, \textbf{74.6}) \\ \hline
Screw(CAD-SD)       & (\textbf{100}, 64.7)  & (\textbf{100}, 73.0)  & (\textbf{100}, \textbf{81.3})  \\ \hline
\end{tabular}}
\end{table}

\begin{table*}[t]
\caption{Quantitative comparisons of image-level detection results with different selected features on the logical anomaly detection benchmarks. The results are represented as (logical anomaly detection AUROC \%, structural anomaly detection AUROC \%) }
\centering
\label{Table8}
\scalebox{1}{
\begin{tabular}{c|cccc}
\hline
Category            & $A$           & $A+Co$         & $A+H$         & $A+Co+H$       \\ \hline
Breakfast Box       & (83.4, 69.3) & (\textbf{95.2}, \textbf{70.3}) & (83.0, 69.6) & (94.5, 70.0) \\
Juice Bottle        & (82.3, 67.0) & (\textbf{90.8}, 80.0) & (85.3, 67.3) & (\textbf{90.8}, \textbf{80.5}) \\
Pushpins            & (80.2, 57.0) & (79.0, \textbf{95.7}) & (88.9, 59.8) & (\textbf{89.0}, 93.8) \\
Screw Bag           & (74.4, 62.0) & (79.0, 62.8) & (75.0, 62.1) & (\textbf{79.9}, \textbf{65.0}) \\
Splicing Connectors & (79.6, 60.3) & (82.4, 61.3) & (80.8, 62.3) & (\textbf{84.3,} \textbf{63.8}) \\ \hline
Average.MVTec LOCO & (80.0, 63.1) & (85.3, 74.0) & (82.6, 64.2) & (\textbf{87.7}, \textbf{74.6}) \\ \hline
Screw(CAD-SD)       & (\textbf{100}, 76.0)  & (\textbf{100}, \textbf{81.8})  & (\textbf{100}, 76.4)  & (\textbf{100}, 81.3)  \\ \hline
\end{tabular}}
\end{table*}

\subsection{Segmentation granularity}
\label{5.2}
\begin{figure}
    \centering
		\includegraphics[width=\columnwidth]{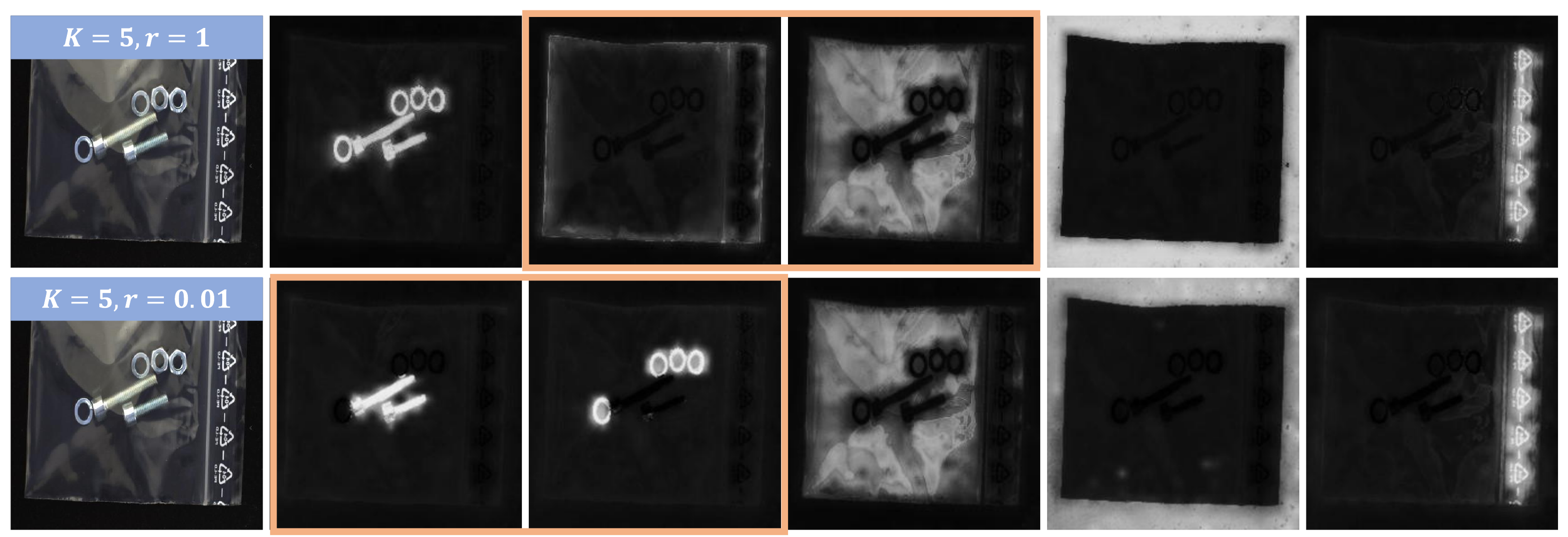}
	\caption{Qualitative comparisons when applying the same $K$ but different corset sampling ratio $r$. }
	\label{FIG:13}
\end{figure}

The segmentation granularity has a significant impact on the use of the model. In ComAD, we first apply corset sampling to the feature maps and then use KMeans to cluster the reserved features. We find that both the corset sampling ratio $r$ and the $K$ of KMeans can affect the segmentation granularity. 
For the corset sampling ratio $r$, under the same $K$ value, a higher $r$ will lead to more fine-grained segmentation of the background region, as the background region usually occupies the majority of the reserved features. Conversely, a lower $r$ can reduce the weight of the background region, thereby allowing the model to focus more on segment foreground components. A typical example is shown in Fig. \ref{FIG:13}, where there is a `screw bag' image with the same $K$ but different $r$. It can be observed that when $r=0.01$, the model has achieved the detailed segmentation on the foreground `screw' and the `washer and the nut', while when $r=1$, the model exhibit over-segmentation of the `bag'. In this scenario, we recommend using a lower value of $r$, which can make the model focus more on the foreground components and also reduce memory consumption.

\begin{figure}
    \centering
		\includegraphics[width=\columnwidth]{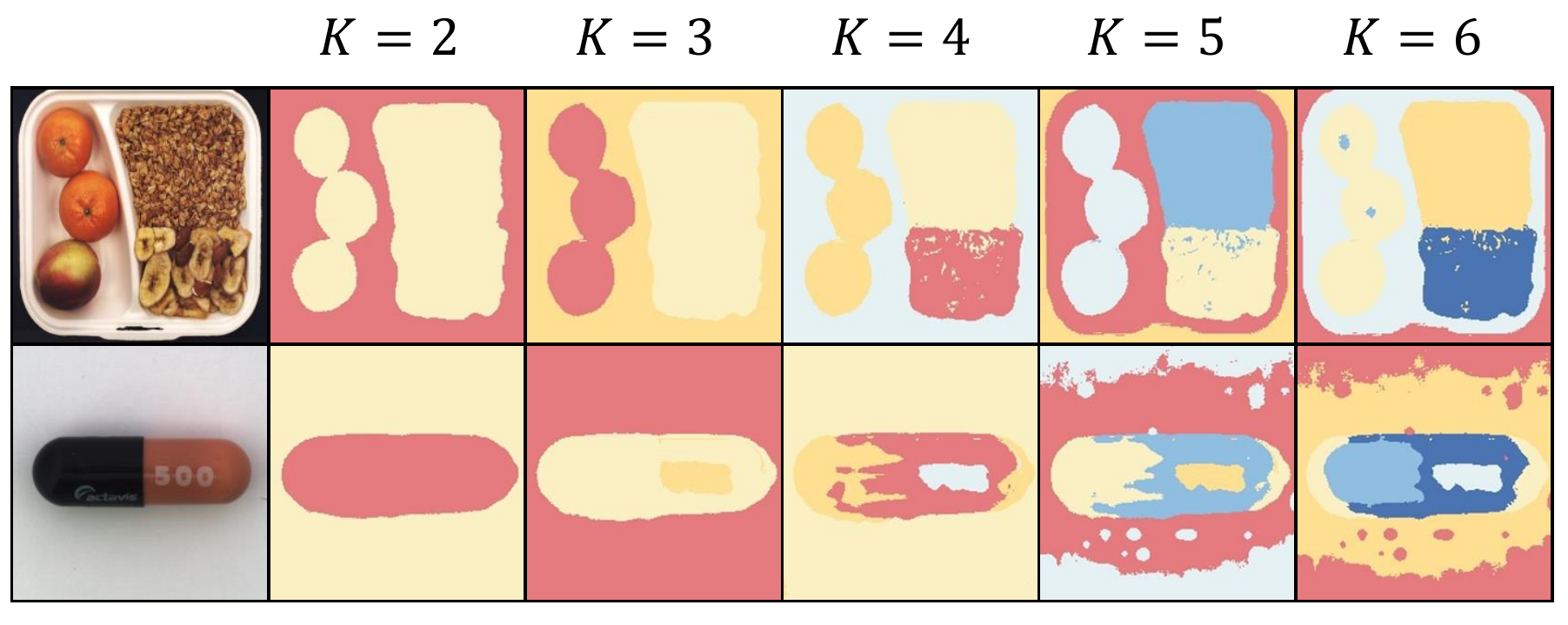}
	\caption{Qualitative comparisons when applying different $K$ values on the multi-component product (`breakfast box') and single-component product (`capsule' from the MVTec AD dataset).}
	\label{FIG:14}
\end{figure}

For the value of $K$ in KMeans, increasing $K$ can generally lead to finer-grained segmentation results in multiple-component products. However, for products with relatively fewer components, this may lead to over-segmentation, introducing meaningless noise. A typical example is shown in Fig. \ref{FIG:14}, where an increasing number of $K$ benefits more detailed segmentation of the `breakfast box' image while harming the segmentation of the `capsule' image. Furthermore, we conduct experiments to verify the impact of different $K$ values on the anomaly detection tasks. We combine our model with PatchCore, which has shown the best performance for previous ensemble detection. The results are shown in Table. \ref{Table5}. We observe that larger $K$ values are more suitable for products with multiple components. Conversely, for single-component products such as the products in the MVTec AD object categories, it's preferable to reduce the $K$ value.  

Besides, we find that no matter how much we increase the value of $K$, the model always can not distinguish between `oranges' and `peaches' in the `breakfast box' category, despite their obvious visual differences. Similar examples include `nuts' and `washers' in the `screw bag' category, and different colored `wires' in the `splicing connector' category. These are the limitations of our model. To further differentiate these components, it may be necessary to introduce additional features.  

\subsection{The influence of CRF}
\label{5.3}
\begin{figure}
    \centering
		\includegraphics[width=\columnwidth]{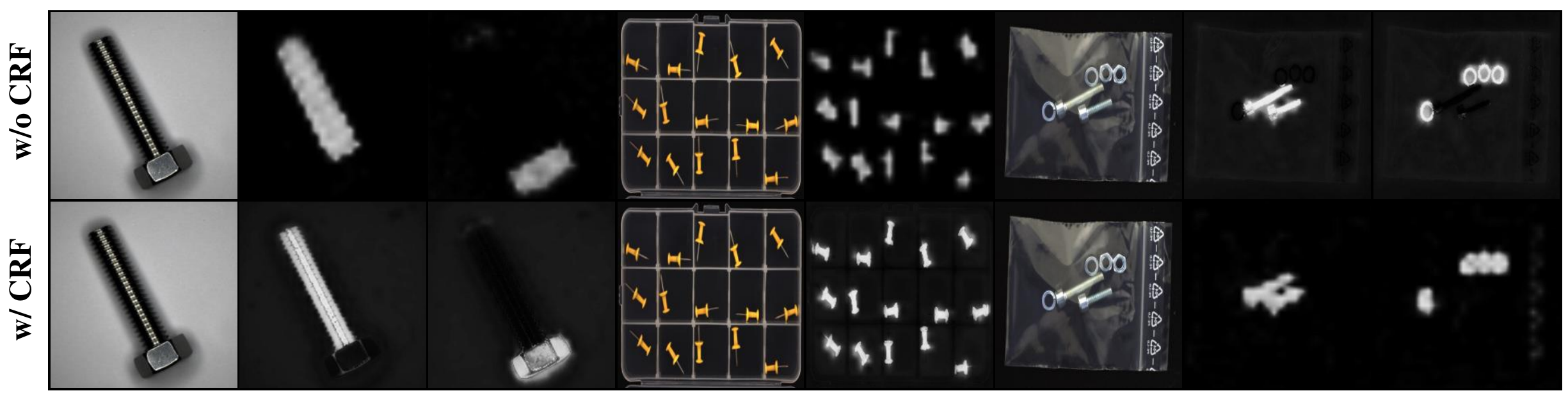}
	\caption{Qualitative comparisons of segmented results with (w/) and without (w/o) CRF.}
	\label{FIG:15}
\end{figure}

CRF is capable to refine the initial segmentation results. We show some qualitative segmentation results with and without CRF in Fig. \ref{FIG:15}. It can be observed that using CRF can result in sharper segmentation results. We further conduct experiments to verify its impact on our anomaly detection tasks. As shown in Table. \ref{Table6}, incorporating CRF generally improves the overall performance but decreases the performance in the `screw bag' category. In this case, although using CRF produces clearer results, the segmentation results without CRF may be more consistent, as in the case of ring-shaped objects. Also, using CRF will cost more computational time. In our configuration, CRF itself takes about 35ms per image, while without CRF, the entire segmentation process takes less than 5ms per image. For some simple cases, such as the logical anomalies in the screw of the CAD-SD dataset, CRF may be unnecessary and can be omitted to improve efficiency.

\subsection{The analysis of the logical anomaly detection model}
\label{5.4}
\textbf{Region extraction}. As previously shown in Fig. \ref{FIG:5}, The leverage of the `argmax' function, OTSU, and our Adaptive scaled OTSU can lead to different region extraction results. Here we verify their impacts on anomaly detection. As shown in Table. \ref{Table7}, we find that using multi-class competition for region extraction (`argmax') is generally suboptimal compared to the binary classification of each component map (`OTSU' and `Adaptive scaled OTSU'). Compared to OTSU, the proposed Adaptive scaled OTSU improves the performance in categories with more segmentation noise, such as the `pushpins' and the `screw bag' categories.

\textbf{Selected region features}. We leverage the region area features ($A$), color features within the region ($Co$), and region connectivity ($H$) to model the metrological features of the components. Here we conduct ablation experiments to analyze their effects, and the results are shown in Table. \ref{Table8}. We find that incorporating color information can significantly improve the overall performance of the model, since the original DINO features may not effectively distinguish components that have similar semantics but different colors. The region connectivity is primarily applicable to components with distinct separations, such as the `pushpins' category. In this scenario, it can serve as an indirect counting approach to detect logical anomalies caused by the wrong number of components.

\subsection{Limitations}
\label{5.5}
Our component segmentation model relies on the DINO pre-trained features, so it's limited by the representation ability of DINO and therefore may require additional post-processing algorithms in specific scenarios. This issue may be addressed by more powerful zero-shot segmentation models \cite{kirillov2023segment} with a trade-off between model complexity and accuracy. As for the model's adjustability, our model is designed to classify the anomalies based on the components they belong to. Therefore, it is not able to classify anomalies based on their own texture features and not applicable to homogeneous texture anomaly detection tasks. For logical anomaly detection, we primarily utilize semantic segmentation results. However, these segmentation maps lack instance-level perception, resulting in suboptimal performance on complex counting problems and instance-pose anomalies. Meanwhile, our selected area and color features may struggle to represent complex situations. In these cases, the selected features need to be expanded. For logical anomaly localization, we can trace back to the anomalous components, but it is challenging to present pixel-level results. For example, when a component is missing, we can identify which component is missing but it is difficult to indicate its intended location in the image.  

\section{Conclusion}
\label{6}
In this paper, we propose to segment the product into multiple components for solving adjustable industrial visual inspection and logical anomaly detection. Specifically, we propose a simple yet effective component segmentation model and an explainable logical anomaly detection model. The proposed framework can achieve more adjustable anomaly detection which helps better meet the customized tuning requirements for practical applications. Additionally, it achieves state-of-the-art performance on logical anomaly detection tasks. Besides, our framework is lightweight and requires minimal training. Overall, our approach differs from previous logical anomaly detection methods that focus on long-range modeling. Therefore, we hope it can bring a new perspective to the challenging logical anomaly detection tasks and becomes more effective with the advancements in more powerful large zero-shot visual perception models.  

\section*{Declaration of Competing Interest}
The authors declare that they have no known competing financial interests or personal relationships that could have appeared to influence the work reported in this paper.

\section*{Acknowledgement}
The calculations were performed by using the HPC Platform at Xi’an Jiaotong University.




\end{document}